%% file: paper.tex
\documentclass[]{taoflowforge}

\usepackage[toc,page,header]{appendix}

\usepackage{mathrsfs}
\usepackage{adjustbox}
\usepackage{multirow}
\usepackage{multirow}
\usepackage{multicol}
\usepackage{tcolorbox}
\usepackage{changepage}
\usepackage{graphicx}
\usepackage{amssymb}
\usepackage{array}
\usepackage{bm}
\usepackage{hyperref}
\usepackage{xspace}
\usepackage{minitoc}
\usepackage{wrapfig}
\usepackage{tikz}
\usepackage{algorithm}
\usepackage{algorithmicx}
\usepackage{algpseudocode}
\usepackage{amsmath,amssymb}
\usetikzlibrary{positioning,calc,arrows.meta,fit}
\newcommand{\shortname}{TaoFlowForge\xspace}

\title{TaoFlowForge: Progressive Native Mesh Generation via Cascaded Flow Matching}

\author{Xianze Fang\textsuperscript{*}}
\author{Qiyuan Feng\textsuperscript{*}}
\author{Dongfang Sun}
\author{Yan Zhang}
\author{Xiuchao Wu}
\author{Jingnan Gao}
\author{\\Jiangjing Lyu\textsuperscript{\dag}}
\author{Chengfei Lyu\textsuperscript{\dag}}
\author{Gang Yu}

\affiliation{Taobao3D Team, Alibaba Group}
\contribution[*]{These authors contributed equally}
\contribution[\dag]{Project Leaders}

\abstract{
    3D content generation technology has significantly advanced the work of designers, as well as the 3D printing and gaming industries.
    However, it remains difficult to produce lightweight, editable, and topologically clean artistic content that is directly production-ready.
    To achieve this, we present \textbf{\shortname}, an artistic mesh foundation model that generates production-ready meshes.
    Specifically, \shortname decomposes the mesh generation process into vertices generation and their connectivity prediction, i.e., edges.
    We formulate vertices generation as a two-stage coarse-to-fine process and incorporate several effective loss functions to further enhance its performance.
    In the connectivity prediction stage, we propose a simple yet effective method for estimating the connectivity affinity between vertices and additionally predict per-vertex normals, which determines the correct orientation of faces.
    Besides, we construct a large-scale dataset combining hand-crafted 3D assets with public high-quality topology datasets. 
    Based on this, a carefully designed data curation pipeline is employed to filter the raw dataset, retaining only high-quality topology data for model training.
    Our model is trained on the combined dataset and tested on both out-of-distribution hand-crafted set of 3D assets and public datasets.
    Under image-conditioned generation, \shortname outperforms autoregressive methods and achieves state-of-the-art results among open-source mesh topology generators.
    We will release all the code and weights together with a portion of our test dataset.
}

\date{\today}
\checkdata[Project Page]{\url{https://alibaba.github.io/Taobao3D/blog/taoflowforge/} }

\begin{document}

\titlefigure{%
  \begin{center}
  \includegraphics[width=1.0\textwidth]{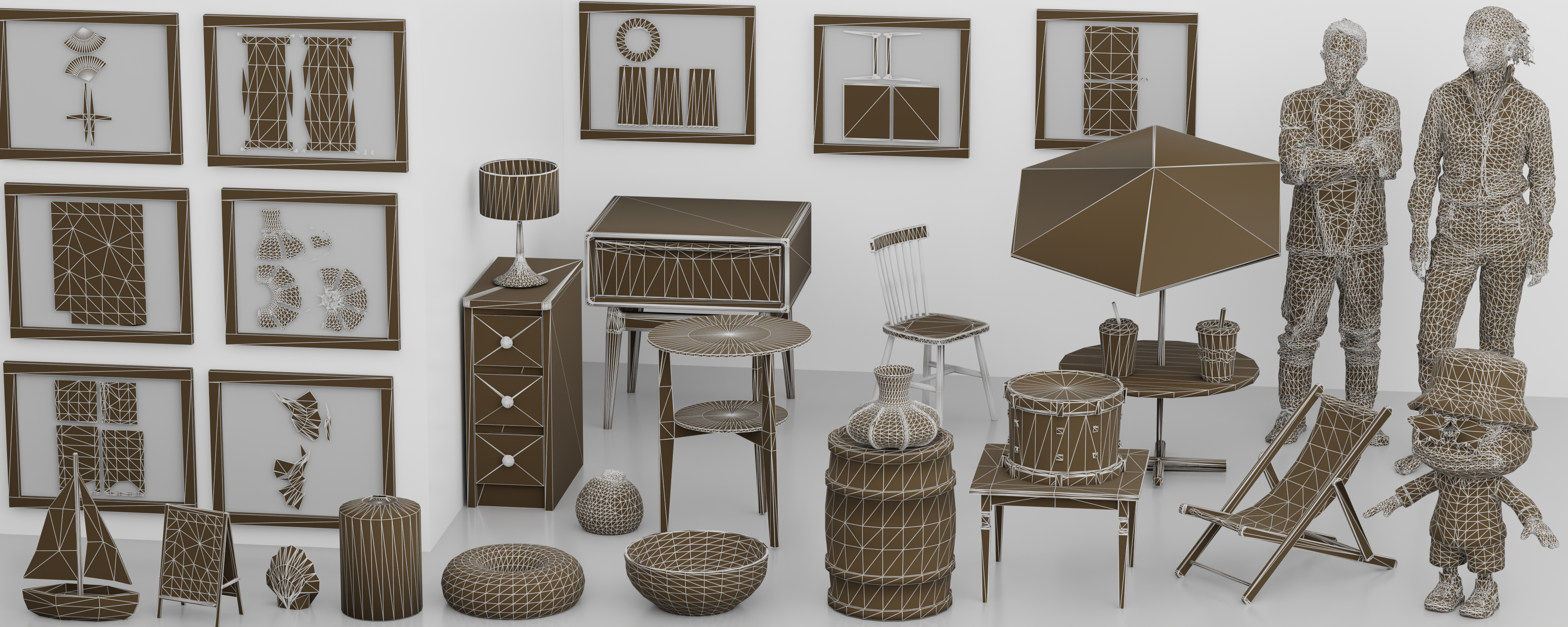}
  \captionof{figure}{Native meshes generated by \shortname across categories (furniture, characters, props), rendered with their predicted topology. The framed images show the UV unwrappings of several meshes in the figure, also revealing the quality of their topology. \textit{Best viewed with zoom-in.}}
  \end{center}
}

\maketitle

\input{sec/1_intro}
\input{sec/related}
\input{sec/2_method}
\input{sec/3_exp}
\input{sec/4_con}

\clearpage
\bibliographystyle{plainnat}
\bibliography{main}

\end{document}

%% file: sec/1_intro.tex
\section{Introduction}
Triangle meshes are the fundamental basis of modern 3D production pipelines, including rigging, animation, and rendering engines. 
Image-conditioned 3D mesh generation models can greatly simplify the workflows for artists and designers in these domains. However, not all generated triangle meshes are ready for industrial deployment.
Models based on SDF 
representations~\cite{zhang20233dshape2vecset,zhang2024clay,zhao2025hunyuan3d,li2025triposg,dora,wu2024direct3d,guo2025hyper3d,xiang2025structured,he2025sparseflex,li2025sparc3d,wu2025direct3d} typically produce meshes with excessively high face counts and topologically irregular structures, which poses three major challenges: (1) UV unwrapping on complex topologies bottlenecks the entire pipeline; (2) heavy storage overhead hinders real-time on-device rendering; and (3) irregular mesh topology complicates downstream rigging and animation.
These issues are especially pronounced on large-scale e-commerce platforms such as Taobao, where 3D assets serve as the interactive content of immersive VR shopping (such as the Vision Pro release of Taobao) and of on-device product display and AR placement on mobile devices.

Recent work has explored the direct generation of mesh topology, bypassing the intermediate SDF representation, in order to produce meshes with a reasonable face count and clean topology.
For example, methods like DeepMesh~\cite{zhao2025deepmesh}, BPT~\cite{weng2025bpt}, and~\cite{nash2020polygen,siddiqui2024meshgpt,chen2024meshanything,tang2024edgerunner,lionar2025treemeshgpt} directly generate the discrete vertex and face sets $(\mathcal{V}, \mathcal{F})$ in a sequential manner. 
These autoregressive methods typically rely on a tokenizer to map meshes into a one-dimensional token space, where generation is formulated as a next-token prediction process. 
The generated token sequence is then converted back into a mesh using a detokenizer.
Such methods are often computationally inefficient at inference time. 
More importantly, encoding meshes as one-dimensional token sequences discards the intrinsic structural information present in native 3D representations, which frequently results in unstable generation performance.
In contrast to these methods, Nexus~\cite{wang2026nexus}, LATO~\cite{zhao2026lato} and
% \fxz{TBD: also cite LATO1} 
LATO.2~\cite{lato2} directly generate 3D primitives with diffusion models~\cite{dit}, achieving higher-quality topology generation. 
However, these methods still suffer from unstable vertex generation quality or the lack of explicit orientation prediction for generated faces.
In addition to the methodological shortcomings of existing research, the shortage of appropriate training data constitutes another major challenge. 3D datasets are inherently limited and scattered across diverse sources, while only a relatively small portion satisfies the quality requirements for clean topology. 
As a result, obtaining sufficient high-quality topological data for model training remains difficult.

To bridge this gap, we propose \shortname, a foundation model for native production-ready 3D content generation.
Concretely, we decompose the pipeline into three stages. 
In the first stage, a latent-space DiT~\cite{dit} generates coarse vertices of an object in a low-resolution canonical space. 
Then, a mixture-of-experts(MoE) model in the original space refines and directly upsamples these primitives to high-fidelity vertex sets. 
The two-stage coarse-to-fine design (we call it \textbf{late-stage refinement}) for vertex generation enables the vertices to maintain a globally correct structure while producing finer geometric details.
Finally, in stage 3, a DiT is used to generate per-vertex features, from which a decoder predicts both the connectivity between vertices and the normal of each vertex, thereby reconstructing the final mesh with correct face normal directions.

Besides the generation architecture, we also construct a large-scale training dataset from Taobao 3D assets and publicly available datasets. 
A rigorous curation pipeline retains only meshes with high-quality topology. 
Training on this diverse dataset enables \shortname to generalise across a broad range of object categories. 
Our evaluations show that \shortname achieves state-of-the-art performance among open-source methods and remains competitive with proprietary commercial methods.

In summary, our core comtributions are as follows:
\begin{itemize}
    \item A novel framework \shortname that generates product-ready 3D meshes through a cascaded late-stage refinement procedure, ensuring structural stability and high fidelity at the same time.
    \item Topology-aware losses designed for vertex generation, including boundary-aware weighted binary cross-entropy to address class imbalance, degree-aware regularization to reinforce structurally important vertices and coplanarity constraints to preserve face structure.
    \item A joint prediction algorithm that simultaneously predicts vertex connectivity and face normals, enabling accurate face recovery with correct orientation and improving topology prediction accuracy.
\end{itemize}

%% file: sec/related.tex
\section{Related Work}
\label{sec:related}

\paragraph{Field-based mesh generation} 
3D content generation has evolved from object-specific optimization with pretrained 2D diffusion priors~\cite{poole2022dreamfusion,wang2023prolificdreamer} to feed-forward reconstruction models~\cite{hong2023lrm,tang2024lgm,yang2024hunyuan3d}.
Field-based methods decode implicit shapes from compact latents, with advances in vector sets~\cite{zhang20233dshape2vecset}, sharp-edge-aware autoencoding~\cite{dora}, triplane diffusion~\cite{wu2024direct3d}, and hybrid representations~\cite{guo2025hyper3d,zhang2024clay,zhao2025hunyuan3d,li2025triposg}.
TRELLIS~\cite{xiang2025structured} introduces sparse spatial structures that concentrate computation near object surfaces, followed by related sparse and structured approaches~\cite{he2025sparseflex,li2025sparc3d,wu2025direct3d,xiang2025native}.
However, the extracted meshes are often dense and topologically unstructured, motivating native-mesh methods that directly generate vertices and faces using autoregressive or diffusion models.

\paragraph{Autoregressive native mesh generation} 
Autoregressive methods directly generate the discrete vertex and face sets in a sequential manner.
\cite{nash2020polygen} generates vertices followed by faces expressed as vertex-index sequences. 
\cite{siddiqui2024meshgpt} generates indices into a learned vocabulary embeddings. 
\cite{chen2024meshanything} uses pointcloud features to condition artist-mesh generation on a target shape. 
\cite{zhao2025deepmesh} further explores preference alignment for autoregressive mesh generation through Direct Preference Optimization (DPO).
To reduce sequence redundancy, \cite{chen2025meshanythingv2} reuses shared edges between consecutive adjacent triangles. 
\cite{tang2024edgerunner}, \cite{weng2025bpt} further reduces the coordinate-token costs by improving the tokenizer.
Complementary to token compression, \cite{hao2024meshtron} combines an hourglass architecture, truncate-sequence training, and sliding-window inference to support up to 64K faces. 
\cite{wang2026face} assigns one transformer token per face, retaining autoregressive coordinate prediction within each face through a CausalMLP head. 
\cite{kim2025fastmesh} restricts autoregressive generation to vertices, then predicts vertex-pair connectivity in parallel
and recovers faces from mutually connected triplets. 
These autoregressive methods share two fundamental limitations. 
Firstly, their sequential decoding makes inference latency scale with sequence length.
Secondly, serializing mesh topology into one-dimensional token sequences discards the intrinsic spatial information in 3D assets, which frequently leads to unstable generation as geometric complexity increases. 

\paragraph{Diffusion-based native mesh generation} 
Diffusion and flow-matching methods update mesh representations within each sampling step. 
\cite{alliegro2023polydiff} applies discrete diffusion to quantized triangle soups, recovering coordinates from categorical noise. \cite{he2025meshcraft} encodes meshes into continuous face-level Variational Autoencoder(VAE) \cite{kingma2022vae} tokens and generates them with a diffusion transformer conditioned on the target face count. 
Both represent meshes through triangle faces rather than a separately generated vertex set and its connectivity.
\cite{li2026meshflow} jointly encodes vertex positions, normals, and adjacency into continuous MeshVAE latents modeled by a diffusion transformer. 
\cite{zhao2026lato} uses a two-stage pipeline to generate coarse structure voxels and predicts edges between vertex pairs using a connection head.
\cite{wang2026nexus} generates vertices through a diffusion model conditioned on preceding octree levels, followed by per-vertex topology latents prediction that recovers edges and faces. 
 These works mark a significant advance over autoregressive methods by generating mesh directly in 3D space, thereby preserving the native spatial structure of meshes and achieving better generation performance. 
 However, it still remains challenges in generating a stable mesh structure and consistent vertex connectivity.
 LATO.2 lacks explicit face orientation prediction, resulting in inconsistent face normals and require post-correction. 
 Nexus relies on hierarchical octree diffusion for vertex generation, where errors accumulate across octree levels, leading to an unstable mesh structure potentially. 
 In this work, we try to address these limitations through our method.

%% file: sec/2_method.tex
\section{Method}

\begin{figure}[t]
    \centering
    \includegraphics[width=\textwidth]{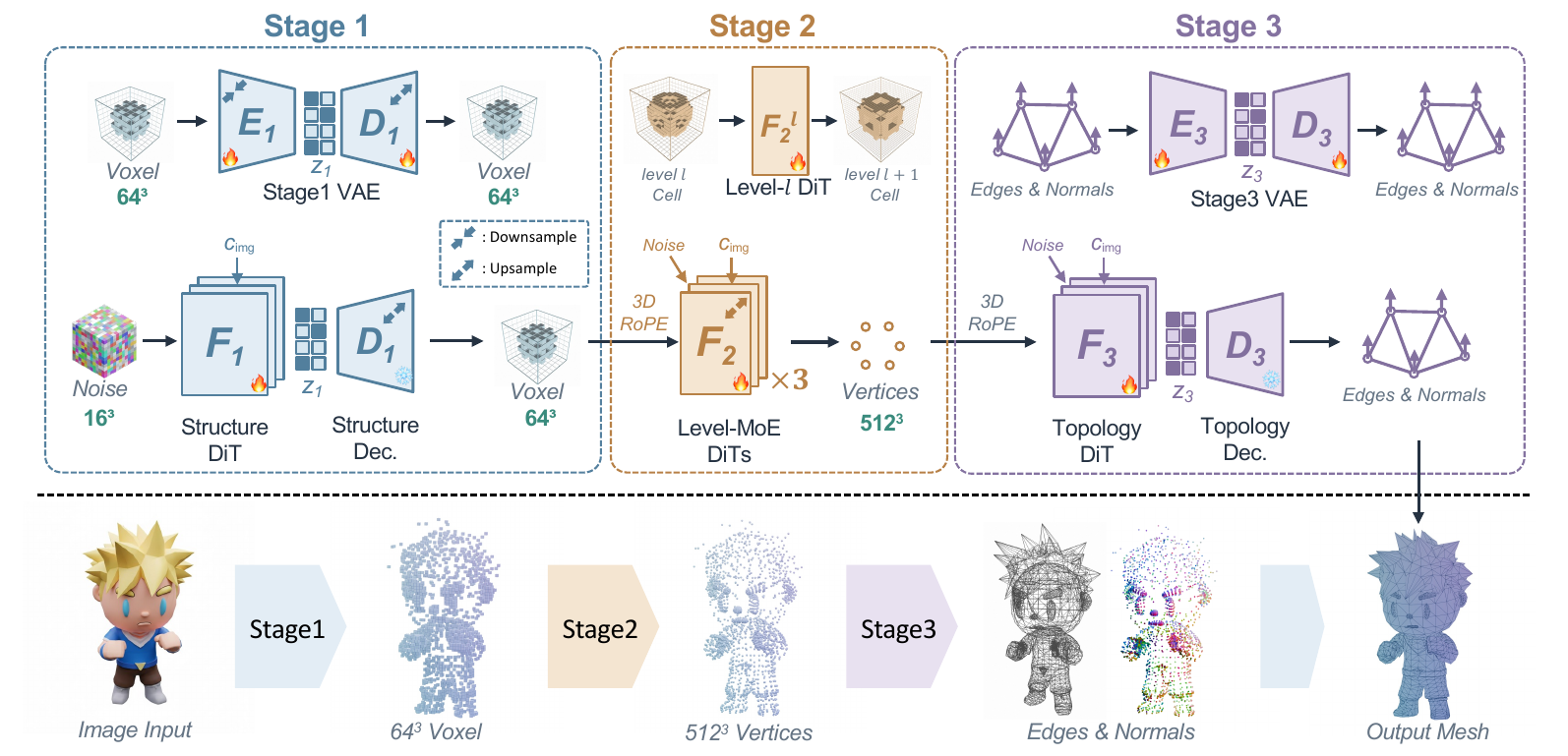}
    % \caption{\sdfc{F should be mentioned in methods.}}
    \caption{Overview of our proposed \shortname: Stage 1: Coarse-quantized vertices are generated at $64^3$ resolution. Stage 2: Vertex positions are progressively refined from $64^3$ to $512^3$. Stage 3: Based on the generated vertices, we predict the final vertex connectivity to form mesh faces.
    Here, $E$ and $D$ denote the encoder and decoder, respectively, and $F$ denotes the flow matching model in each stage, conditioned on image features $c_{\mathrm{img}}$.
    }
    \label{fig:pipeline}
\end{figure}

As a preliminary, a triangle mesh is conventionally represented as a set of vertices $\mathcal{V}$ and faces $\mathcal{F}$. 
In our implementation, each face is defined by three edges in the set of edges $\mathcal{E}$ with a normal orientation $\mathbf{n}$.
For topology generation, the vertices $\mathcal{V}$ are quantized to grid center points at a given resolution, which reduces the search space and improves generation quality.

Based on the $(\mathcal{V}, \mathcal{E}, \mathcal{F})$ data structure, 
% \xchao{grammar mis.}it 
\shortname{} generates product-ready 3D content through three cascaded image-conditioned stages (Fig.~\ref{fig:pipeline}).
Specifically, stage~1 adapts a diffusion generation model to produce a coarse $64^{3}$ occupancy voxel as the coarse-quantized vertex positions.
Then, stage~2 operates directly in the generated $64^{3}$ occupancy voxel, applying flow matching (FM) in normalized target spaces, where three level-specific experts are used to progressively refine the field to $512^{3}$ resolution in a late-stage manner, yielding the final vertices $\hat{\mathcal{V}}$.
Finally, stage~3 generates the per-vertex features through a diffusion model in the latent space.
A topology decoder then predicts both the edge set $\hat{\mathcal{E}}$ and face normals, yielding the final faces $\hat{\mathcal{F}}$ with correct orientations.
All three stages are conditioned on image features $c_{img}$ extracted by DINOv3~\cite{simeoni2025dinov3}.

\subsection{Stage 1: Prior Adaptation for Coarse Structure}
\label{subsec:stage1}

Given an input image, \shortname{} first generates coarse-quantized vertex positions for the target. 
Concretely, let $\mathbf{O}_{64}\in\{0,1\}^{1\times64\times64\times64}$ denote the coarse occupancy voxel grid at $64^{3}$ resolution, which is then compressed by a VAE encoder $E_{1}$ to $\mathbf{z}_{1}\in\mathbb{R}^{d_1\times16\times16\times16}$. 
% Firstly,
First,
% \xchao{better to use first,not firstly} 
the sparse-structure encoder $E_{1}$ predicts the mean $\boldsymbol{\mu}_{1}$ and standard deviation $\boldsymbol{\sigma}_{1}$ of a diagonal Gaussian, defining the approximate posterior $q_{1}(\mathbf{z_1}\mid\mathbf{O}_{64})$ over the latent $\mathbf{z_1}$.
Correspondingly, the mirrored decoder $D_{1}$ maps $\mathbf{z_1}$ back to a $64^{3}$ field of occupancy logits.
The compress process in encoder and mapping process in decoder correspond to the downsampling and upsampling process in Figure.~\ref{fig:pipeline}, respectively.
The encoder process and the approximate posterior can be formulated as
\begin{equation}
    (\boldsymbol{\mu}_{1},\boldsymbol{\sigma}_{1})
    = E_{1}(\mathbf{O}_{64}),
    \qquad
    q_{1}(\mathbf{z_1}\mid\mathbf{O}_{64})
    = \mathcal{N}\!\left(\boldsymbol{\mu}_{1},
    \operatorname{diag}(\boldsymbol{\sigma}_{1}^{2})\right).
\end{equation}

Based on the latent $\mathbf{z_1}$ compressed by the VAE, a flow matching(FM) model is trained to accomplish the generation of coarse-quantized vertices from $c_{img}$. The DiT is trained to predict the deterministic posterior mean $\boldsymbol{\mu}_{1}(\mathbf{O}_{64})$ from the encoder $E_{1}$, standardized channel-wise.
With the sampled noise $\mathbf{z}_{1}^{0}\sim\mathcal{N}(0,I)$ and timestep $t\in(0,1)$, we use the linear interpolation path $\mathbf{z}_{1}^{t}=(1-t)\mathbf{z}_{1}^{0}+t\mathbf{z}_{1}$. Its conditional velocity is constant along the path, giving the objective.
\begin{equation}
    \mathcal{L}_{\mathrm{S1\_DiT}}
    = \mathbb{E}\!\left[
    \left\|v_{\theta}(\mathbf{z}_{1}^{t},t,c_{\mathrm{img}})
    -(\mathbf{z}_{1}-\mathbf{z}_{1}^{0})\right\|_{2}^{2}
    \right].
    \label{eq:s1}
\end{equation}
At inference, the learned velocity field is initialized from Gaussian noise. The resulting latent is then denormalized and decoded by the decoder $D_{1}$. Thresholding the decoder logits yields the coarse $64^{3}$ occupancy voxel, serving as the coarse-quantized vertex positions.
Since our stage 1 adopts the same architecture as the first stage of Trellis.2~\cite{xiang2025native}, we are able to initialize it from the corresponding pretrained checkpoint and fine-tune it directly. This enables us to effectively leverage its learned prior for predicting the coarse object structure from images, while also substantially accelerating our training process.

\subsection{Stage 2: Normalized Raw-Space Refinement with Level-MoE}
\label{subsec:stage2}

Based on the coarse-quantized vertices, stage~2 aims to refine vertex positions through a multi-resolution refiner. Unlike stage~1, it operates directly in the uncompressed occupancy space without using a VAE. 
% For each active parent cell $g$, the model predicts a binary vector $\mathbf{y}^{(g)}\in{\{0,1\}}^{8}$ indicating which of its eight child cells are occupied. 
For each active parent cell $g$, the model incorporates its spatial location through Rotary Positional Encoding (RoPE) and predicts a binary vector $\mathbf{y}^{(g)}\in{\{0,1\}}^{8}$ to indicate which of the eight child cells are occupied.
\paragraph{Normalized Raw-Space.}
As the full child-occupancy target $\mathbf{y}^R$ at resolution $R$ is binary, its statistics shift sharply with resolution, placing the raw target at a very different scale from the Gaussian FM source.
Let $\rho_{R}=\mathbb{E}[y^{(g)}]\in(0,1)$ be the occupied fraction at resolution $R$, estimated once over the training set. We whiten each level with its own Bernoulli statistics,
\begin{equation}
    \widetilde{\mathbf{y}}^{R}
    = \frac{\mathbf{y}^{R}-\rho_R}
    {\sqrt{\rho_R(1-\rho_R)}},
    \label{eq:s2norm}
\end{equation}
where the mean $\rho_{R}$ centres the target and the Bernoulli standard deviation $\sqrt{\rho_{R}(1-\rho_{R})}$ rescales it, so that $\widetilde{\mathbf{y}}^{(\ell)}$ has zero mean and unit variance and shares the first two moments of the standard Gaussian source. With a source draw $\boldsymbol{\epsilon}\sim\mathcal{N}(0,I)$ we form the linear FM path
\begin{equation}
    \mathbf{x}_{t}^{R}
    = (1-t)\boldsymbol{\epsilon}
    + t\widetilde{\mathbf{y}}^R,
    \qquad \boldsymbol{\epsilon}\sim\mathcal{N}(0,I),
    \label{eq:s2path}
\end{equation}
whose timestep $t\in(0,1)$ interpolates from pure noise at $t=0$ to the whitened target at $t=1$. The resolution $R$ expert $f_{R}$ consumes the interpolated state $\mathbf{x}_{t}^R$, the time $t$, and the DINOv3 image condition $c_{\mathrm{img}}$, and outputs eight raw occupancy logits $\mathbf{a}_R=f_R(\mathbf{x}_{t}^R,t,c_{\mathrm{img}})\in\mathbb{R}^{8}$, one per child cell; their sigmoid $\hat{\mathbf{y}}^R=\sigma(\mathbf{a}_R)$ gives the predicted child-occupancy probabilities. At inference we whiten this prediction with Eq.~\eqref{eq:s2norm} and integrate, in normalized space, the endpoint-induced velocity
\begin{equation}
    v_{\theta_{R}}(\mathbf{x}_{t}^{R},t)
    = \frac{\widetilde{\hat{\mathbf{y}}}^{R}-\mathbf{x}_{t}^{R}}{1-t},
    \qquad t<1,
    \label{eq:s2velocity}
\end{equation}
where $\widetilde{\hat{\mathbf{y}}}^{R}$ 
% \sdfc{You mean $\widetilde{\hat{\mathbf{y}}}^{R}$}\fqyc{done.}
is the whitened prediction and the guard $t<1$ avoids the endpoint singularity; the integrated state is mapped back to occupancy space and thresholded to recover the occupied children.

\paragraph{Level-indexed mixture of experts.}
Within the hierarchical DiT framework, each refinement level increases the vertex resolution by a factor of two.
As the resolution increases, the occupied fraction $\rho_R$ decreases monotonically, yielding progressively sparser and statistically distinct occupancy distributions across levels. 
We therefore assign a dedicated expert to each resolution, allowing it to specialize in its own occupancy statistics. 
The experts operate in cascade order, with occupied cells output from one level becoming the input for the next, as shown in Figure.~\ref{fig:pipeline}.
The final level expert provide the refined $512^{3}$ vertex coordinates $\hat{\mathcal{V}}$.
\begin{figure}[t]
    \centering
    \includegraphics[width=\textwidth]{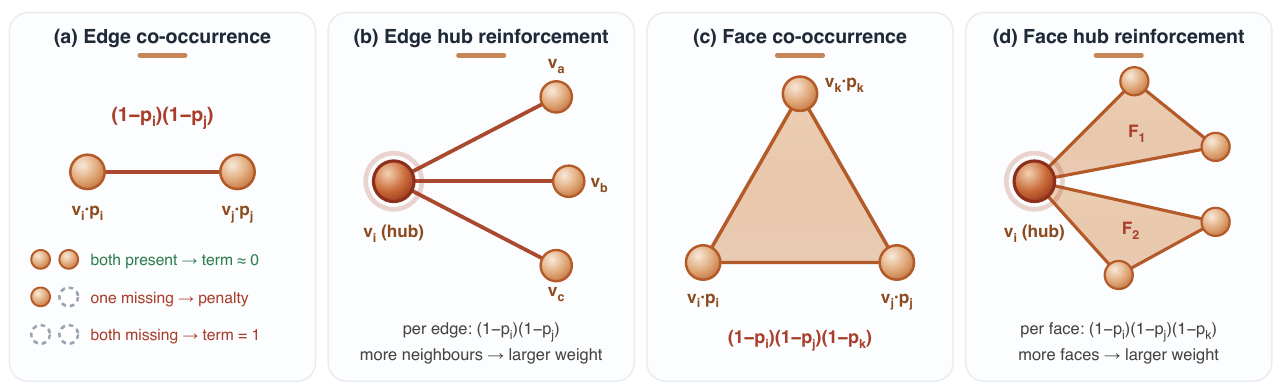}
    \caption{\textbf{Topology-aware co-occurrence priors of stage~2.} Two soft-OR co-occupancy penalties complement per-cell cross-entropy with mesh connectivity constraints. Edge and face priors activate only when all connected vertices are jointly predicted empty, reinforcing structurally important high-degree vertices through accumulated constraints.}
    \label{fig:s2loss}
\end{figure}

\paragraph{Training objectives.}
Every expert is supervised by comparing its predicted child-occupancy probabilities against the binary target $\mathbf{y}^{R}$. 
The primary term is a weighted binary cross-entropy that addresses the strong imbalance quantified above. 
We define $p_c = \sigma(a_c)$ as the predicted occupancy of child cell $c$, with $y_c \in \{0,1\}$ as its ground-truth label. The binary training objective can be defined as
\begin{equation}
    \mathcal{L}_{\mathrm{bce}}
    = \frac{\sum_{c} w_{c}\,\mathrm{ce}_{c}}{\sum_{c} w_{c}},
    \qquad
    \mathrm{ce}_{c} = -\big[\alpha\,y_{c}\log p_{c}
    + (1-y_{c})\log(1-p_{c})\big],
    \label{eq:s2bce}
\end{equation}
where the positive-class weight $\alpha>1$ raises recall on the rare occupied cells, and the spatial weight $w_{c}$ reshapes the negatives: $w_{c}=1$ on occupied cells, $w_{c}=w_{\mathrm{near}}$ on empty cells lying within a radius $r$ of some occupied cell, which is the thin surface band where false positives concentrate, and $w_{c}=w_{\mathrm{far}}$ on the remaining bulk-empty cells, with $w_{\mathrm{near}}>w_{\mathrm{far}}$. 

Because this term scores every cell independently, it constrains neither vertex connectivity nor agreement across a shared edge or face, so we add two geometry-aware priors on the same probabilities (Figure.~\ref{fig:s2loss}). 
Let $\mathcal{E}$ collect the child-cell pairs containing the two
endpoints of each ground-truth edge, and $\mathcal{F}$ collect
the child-cell triples containing the three vertices of each
ground-truth face.
The edge endpoint prior
\begin{equation}
    \mathcal{L}_{\mathrm{edge}}
    = \frac{1}{|\mathcal{E}|}
    \sum_{(i,j)\in\mathcal{E}} (1-p_{i})(1-p_{j})
    \label{eq:s2edge}
\end{equation}
is a soft-OR co-occupancy penalty: it vanishes as soon as either endpoint is confidently occupied and grows only when both endpoints of a true edge are jointly predicted empty, injecting mesh connectivity as an occupancy prior over the per-cell independent cross-entropy, as shown in the first panel of Figure~\ref{fig:s2loss}. 
Furthermore, the objective can be rewritten as:
\begin{equation}
    \mathcal{L}_{\mathrm{edge}}
    = \frac{1}{|\mathcal{E}|}
    \sum_{i\in\mathcal{V}} (\sum_{j \in N(i)}(1-p_{j}))(1-p_{i})
    \label{eq:s2edge_update}
\end{equation}
where $N(i)$ represents the neighborhood of vertex $i$. In this formulation, vertices with more neighbors are assigned larger weights, enhancing the supervision of key points as shown in the second panel of Figure.~\ref{fig:s2loss}.
The face prior carries the same soft-OR logic to the three corners of a ground-truth face, as shown in the last two panel of Figure.~\ref{fig:s2loss},
\begin{align*}
    \mathcal{L}_{\mathrm{face}}
    &= \frac{1}{|\mathcal{F}|}
    \sum_{(i,j,k)\in\mathcal{F}} (1-p_{i})(1-p_{j})(1-p_{k}),\\
    &= \frac{1}{|\mathcal{F}|}
    \sum_{i\in \mathcal{V}} ({\sum_{\{(j,k)|(i,j,k)\in\mathcal{F}\}}}(1-p_j)(1-p_k))(1-p_i)
    \label{eq:s2face}
\end{align*}
In summary, the full stage~2 objective, in expectation over resolution $R$ and flow time $t$, combines the weighted cross-entropy with the three priors,
\begin{equation}
    \mathcal{L}_{\mathrm{S2}}
    = \mathbb{E}_{R,t}\!\left[
    \mathcal{L}_{\mathrm{bce}}
    + \lambda_{\mathrm{e}}\mathcal{L}_{\mathrm{edge}}
    + \lambda_{\mathrm{f}}\mathcal{L}_{\mathrm{face}}
    \right],
    \label{eq:s2}
\end{equation}
whose per-term weights $\lambda_{\mathrm{e}},\lambda_{\mathrm{f}}\ge 0$ are set per resolution to balance the priors against the cross-entropy.

\subsection{Stage 3: Per-Vertex Topology Generation}
\label{subsec:stage3}

Given the vertex set $\hat{\mathcal{V}}$ from stage~2, stage~3 directly predicts vertex connectivity (i.e. edges $\mathcal{E}$) and per-vertex normal orientations as shown in Figure.~\ref{fig:pipeline}.
It combines a topology VAE with a latent flow matching model. 
The VAE learns a continuous feature for each vertex, with vertex connectivity and vertex normal encoded.
During the DiT inference stage, the vertex coordinates associated with voxel centers quantized at a certain resolution in stage~2 are incorporated into the latent features using rotary positional encoding (RoPE).

\paragraph{Topology VAE.}
Unlike the occupancy VAE inherited in stage~1, this VAE performs no spatial compression: it attaches exactly one latent code to each vertex, so the number of latents always equals the vertex count $V=|\hat{\mathcal{V}}|$. 
It encodes two key pieces of information: how the vertex connects to others, as well as the surface normal at that vertex. 
The vertex coordinates, already determined by stage~2, are not re-encoded.

The connectivity head determines whether an edge exists between each pair of vertices.
A naive design would compare the two endpoint features $(\mathbf{f}_{i}, \mathbf{f}_{j})$ by a symmetric similarity, such as the inner product $\langle W\mathbf{f}_{i}, W\mathbf{f}_{j}\rangle$ under a single shared projection. 
However, this inner product inside one embedding space favours \emph{transitive} connectivity: once $A$ is similar to $B$ and $B$ to $C$, the score of $(A,C)$ is pushed up as well, which can produce numerous false connections in mesh adjacency prediction. 
% Mesh adjacency violates this property, since two edges $(A,B)$ and $(B,C)$ routinely share a common vertex $B$ while $A$ and $C$ stay unconnected, so a similarity head hallucinates the spurious edge $(A,C)$ and produces over-connected, non-manifold neighbourhoods. 
% Because mesh adjacency is inherently non-transitive, modeling it as a transitive relation can produce numerous false connections.
We instead project each vertex into two \emph{distinct} subspaces before matching. 
Writing $\mathbf{h}_{i}\in\mathbb{R}^{d}$ for the feature of vertex $i$, a shared edge MLP maps it to $\mathbf{f}_{i}=g_{e}(\mathbf{h}_{i})\in\mathbb{R}^{d_{e}}$, and two independent bias-free projections form source $W_s$ and destination $W_d$ embeddings,
\begin{equation}
    \mathbf{s}_{i}=W_{s}\mathbf{f}_{i},
    \qquad
    \mathbf{d}_{i}=W_{d}\mathbf{f}_{i},
    \qquad W_{s},W_{d}\in\mathbb{R}^{d_{e}\times d_{e}}.
    \label{eq:edgefeature}
\end{equation}
% Stacking them row-wise as $S$ and $D$ gives the directed score matrix $Q=SD^{\top}+b\mathbf{1}\mathbf{1}^{\top}$ with a learned scalar bias $b$, and because mesh edges are undirected we symmetrise it, so the pair logit is
Because mesh edges are undirected, we symmetrise by:
\begin{equation}
    a_{ij}
    = \frac{Q_{ij}+Q_{ji}}{2}
    = \frac{\mathbf{s}_{i}^{\top}\mathbf{d}_{j}
    +\mathbf{s}_{j}^{\top}\mathbf{d}_{i}}{2}+b.
    \label{eq:bilinear-edge}
\end{equation}
$a_{ij}=a_{ji}$ is the logit judging whether there is an edge between vertex $i$ and $j$.
The complete $V\times V$ logit matrix is produced in parallel by batched matrix multiplication.
% self-connections and padded pairs are masked, edge probabilities are $p_{ij}=\sigma(a_{ij})$, and 
At training, the connectivity term $\mathcal{L}_{\mathrm{edge}}$ is a class-balanced binary cross-entropy over the valid upper-triangular logits. 
At inference, thresholding these probabilities defines the edge set $\hat{\mathcal{E}}$.
% \begin{equation}
%     \hat{\mathcal{E}}
%     = \{(i,j): i<j,\ p_{ij}\geq\tau_{e}\}.
%     \label{eq:topoedge}
% \end{equation}
Triangular faces are then recovered as the closed three-cycles of this edge graph,
\begin{equation}
    \hat{\mathcal{F}}
    = \{(i,j,k): i<j<k,\ (i,j),(j,k),(k,i)\in\hat{\mathcal{E}}\}.
    \label{eq:clique}
\end{equation}

The recovered edge graph fixes \emph{which} triangles exist but leaves their orientation undefined.
Unlike some methods such as LATO.2~\cite{lato2} adapting post-process to recover face orientation, we instead let the network emit orientation directly. 
Specifically, we use a normal head in decoder which runs in parallel and regresses a unit normal $\hat{\mathbf{n}}_{i}$ for each vertex from the same feature $\mathbf{f}_{i}$, supervised by a cosine term $\mathcal{L}_{\mathrm{normal}}$ against the ground-truth vertex normals.
For a recovered face $(i,j,k)$ with positions $\mathbf{v}_{i},\mathbf{v}_{j},\mathbf{v}_{k}$ and geometric normal $\mathbf{g}_{ijk}=(\mathbf{v}_{j}-\mathbf{v}_{i})\times(\mathbf{v}_{k}-\mathbf{v}_{i})$, the winding is kept when it agrees with the predicted vertex normals and two vertices are swapped otherwise,
\begin{equation}
    (i,j,k)\ \mapsto\quad
    \begin{cases}
        (i,j,k), & \mathbf{g}_{ijk}\cdot(\hat{\mathbf{n}}_{i}+\hat{\mathbf{n}}_{j}+\hat{\mathbf{n}}_{k})\geq 0,\\[2pt]
        (i,k,j), & \text{otherwise.}
    \end{cases}
    \label{eq:orient}
\end{equation}

The two heads share the decoder and are optimised together with a Gaussian regulariser $\mathcal{L}_{\mathrm{KL}}$ on the per-vertex posteriors, giving the topology VAE objective
\begin{equation}
    \mathcal{L}_{\mathrm{S3\_VAE}}
    = \mathcal{L}_{\mathrm{edge}}
    + \lambda_{n}\mathcal{L}_{\mathrm{normal}}
    % + \beta_{t}\mathcal{L}_{\mathrm{KL}}.
    + \lambda_{r}\mathcal{L}_{\mathrm{KL}},
    \label{eq:topovae}
\end{equation}
in which per-term weights $\lambda_{n}, \lambda_{r} \geq 0$ are set to balance the losses.

\paragraph{Position-aware latent FM.}
The per-vertex latent target $\mathbf{z}_{3}$ is standardised with dataset statistics and paired with Gaussian noise $\mathbf{z}^{0}_{3}$. 
Vertex coordinates from stage~2, quantized at the final resolution level, are injected into every DiT block via three-dimensional RoPE, where each token is assigned the coordinate of its corresponding vertex.
The DiT therefore predicts one topology feature per known vertex while retaining its native spatial location. 
Conditioned on the DINOv3 image feature $c_{img}$, it is trained with the velocity objective
\begin{equation}
    \mathcal{L}_{\mathrm{S3\_{DiT}}}
    = \mathbb{E}\!\left[
    \left\|v_{\psi}(\mathbf{z}^{t}_{3},t,\hat{\mathcal{V}},c_{\mathrm{img}})
    -(\mathbf{z}_{3}-\mathbf{z}^{0}_{3})\right\|_{2}^{2}
    \right],
    \qquad
    \mathbf{z}^{t}_{3}=(1-t)\mathbf{z}^{0}_{3}+t\mathbf{z}_{3}.
    \label{eq:topoflow}
\end{equation}
The generated latents are denormalized before topology decoding.

\subsection{Training Data}
\label{subsec:data}

\paragraph{Data sources.}
\shortname is trained on a large and intentionally heterogeneous mesh dataset drawn from three complementary sources: 
1. Taobao in-house 3D dataset, dominated by everyday indoor commodities and character models.
2. Manually crafted in-house dataset, which densifies the categories our catalogue covers only sparsely. 
3. Public open-source dataset, including Objaverse~\cite{objaverse} and TexVerse~\cite{texverse}.
The scale and diversity of our training dataset are essential for topological generalization and explain why \shortname transfers well to in-the-wild product imagery.
For every asset, we render a full $360^{\circ}$ ring of surrounding views with a physically based renderer, providing the paired image-mesh supervision for training. 
Although most of these meshes already have clean, well-structured topology, the raw collection still contains many low-quality, noisy samples that must be filtered out before they are usable for topology supervision.

\paragraph{Data curation.}
A carefully designed multi-stage curation pipeline is adopted to filter the raw dataset.
We first bucket all assets by face count and sample the buckets as evenly as possible during training, so that neither low-poly nor dense meshes dominate the gradient and the model observes the full spectrum of mesh complexity.
We then screen the entire dataset with a suite of classical topology statistics, including the fraction of non-manifold faces, the distribution of vertex degrees (the number of edges incident to each vertex), and the spatial uniformity of the vertex layout, keeping only assets whose connectivity is regular enough to supply a learnable edge-flow signal.
Beyond these hand-crafted criteria, we further build a VLM-based agent that inspects the rendered views of each surviving asset and prunes the residual low-quality cases that the numeric filters overlook, progressively refining the corpus down to the artist-grade topology on which \shortname is trained. The final curated dataset consists of roughly 800,000 meshes.

%% file: sec/3_exp.tex
\section{Experiments}

\subsection{Implementation Details}
\label{subsec:exp_setup}

\paragraph{Test Set}
For evaluation we use the public Toys4K~\cite{stojanov2021toys4k} test split together with TE-388, a dataset of $388$ manually crafted examples spanning indoor, outdoor, and character categories, which we will release later.

\paragraph{Architecture details.} Stage~1 couples an occupancy VAE (3D convolutional encoder-decoder compressing $64^3 \to 16^3{\times}8$) with a 30-block latent flow DiT ($\approx$1.3B parameters) that cross-attends to frozen DINOv3 tokens. Stage~2 operates in raw occupancy space without a VAE, employing three parameter-independent 28-block transformer experts for the $64^3{\to}128^3$, $128^3{\to}256^3$, and $256^3{\to}512^3$ transitions, with targets standardized per level. Stage~3 pairs a per-vertex topology VAE (VecSet-style encoder pooling into 2048 latent tokens, bilinear decoder predicting pairwise connectivity and normals) with a 32-layer latent flow DiT that injects vertex coordinates via voxel RoPE. All stages cross-attend to DINOv3 tokens for image conditioning.

\paragraph{Evaluation metrics.}
We assess geometry with two surface distances, semantics with two image-shape alignment scores, and rendering fidelity with two Fr\'echet feature distances.
\emph{Chamfer Distance} (CD) is computed between dense point samples of the generated and ground-truth surfaces and reflects mean surface deviation.\emph{Hausdorff Distance} (HD) is taken on the same point samples as the maximum of the two directional worst-case distances.\emph{ULIP-I}~\cite{xue2023ulip} measures whether the generated shape matches the conditioning image in a joint embedding space. \emph{Uni3D-I}~\cite{zhou2024uni3d} follows the same image-to-shape alignment protocol with $10{,}000$ points and the larger Uni3D point encoder aligned to an EVA-CLIP image tower; higher is better. \emph{FD-Inception}~\cite{fid} is a distribution-level Fr\'echet distance over multi-view renders of the generated and reference meshes under a shared untextured shading protocol. \emph{FD-DINOv2} applies the same Fr\'echet distance to features from a DINOv2 ViT-L/14~\cite{oquab2023dinov2} encoder ($1024$-dimensional class token).

\subsection{Image Conditioned Mesh Generation}
\label{subsec:exp_cond}

\shortname is conditioned on a single product image: patch tokens from a frozen DINOv3~\cite{simeoni2025dinov3} backbone are routed into every DiT block through cross-attention.
This image-only interface runs no 3D encoder at inference and matches our production deployment, and we use it for every result reported in this section.

\paragraph{Baselines.}
We compare \shortname against two representative open-source image-conditioned mesh generators: LATO.2~\cite{lato2}, a diffusion-based generator, and EdgeRunner~\cite{tang2024edgerunner}, an autoregressive generator. These two cover the dominant paradigms for native mesh generation, and both take the same single-image input as \shortname, so the comparison is matched in input information. We evaluate on the public Toys4K~\cite{stojanov2021toys4k} benchmark and on TE-388, our proposed test set of $338$ manually crafted meshes spanning indoor, outdoor, and character categories. For each baseline we use the publicly released checkpoint and follow its recommended inference protocol.

\paragraph{Quantitative comparison.}
Table.~\ref{tab:point_cond} reports all six metrics defined in Sec.~\ref{subsec:exp_setup} on Toys4K and TE-388: Chamfer (CD) and Hausdorff (HD) distances for geometric accuracy, ULIP-I and Uni3D-I for image-to-shape semantic alignment, and FD-Inception and FD-DINOv2 for distribution-level rendering fidelity. All three methods are evaluated under an identical single-image protocol, so they differ only in their generative formulation.

\begin{figure}[t]
    \centering
    \includegraphics[width=\textwidth]{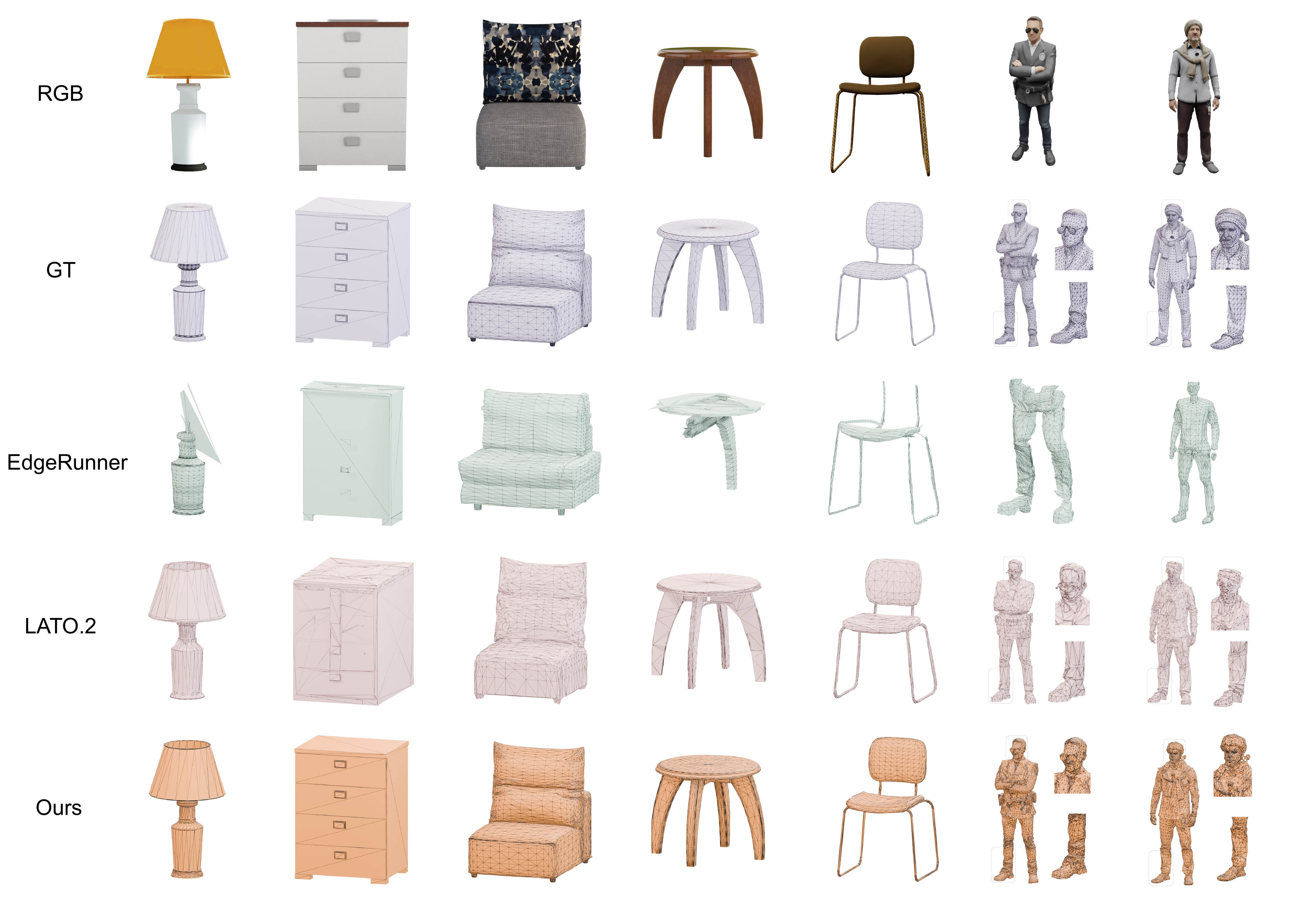}
    % \caption{TO BE FILLED.}
    \caption{Qualitative comparison with open-source polygon-based 3D mesh generation models. 
    Given an image as input, our model generates the finest and most complete 3D meshes.
    }
    \label{fig:main_compare}
\end{figure}

\begin{figure}[t]
    \centering
    \includegraphics[width=\textwidth]{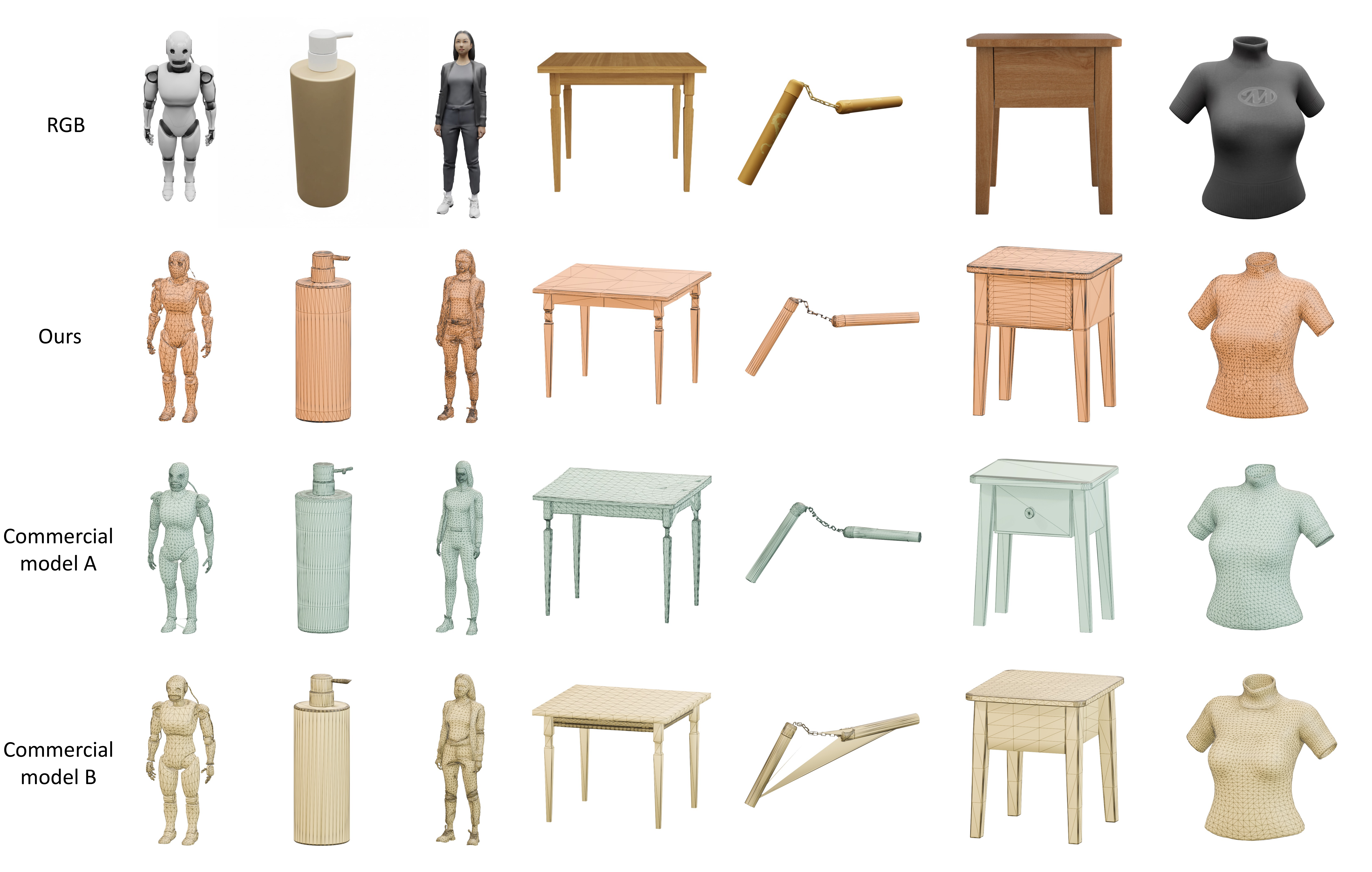}
    \caption{Qualitative comparison with commercial polygon-based 3D mesh generation models. 
    Given an image, our model generates more detailed and stable meshes with better alignment.
    }
    \label{fig:second_compare}
\end{figure}

\begin{table}[h]
    \centering
    \small
    \caption{Quantitative comparison on Toys4K and TE-388 under single-image conditioning. 
    }
    \label{tab:point_cond}
    \begin{tabular}{lcccccc}
        \toprule
        Method & CD $\downarrow$ & HD $\downarrow$ & ULIP-I $\uparrow$ & Uni3D-I $\uparrow$ & FD-Incep. $\downarrow$ & FD-DINOv2 $\downarrow$ \\
        \midrule
        \multicolumn{7}{l}{\emph{Toys4K}} \\
        LATO.2~\cite{lato2}                   & 0.1553 & 0.3663 & 0.1674 & 0.3330 & 78.9649 & 616.4222 \\
        EdgeRunner~\cite{tang2024edgerunner} & 0.3264 & 0.6145 & 0.1514 & 0.2347 & 30.9466 & 339.7754 \\
        \shortname (Ours)                    & \bfseries{0.0655} & \bfseries{0.2451} & \bfseries{0.1923} & \bfseries{0.3509} & \bfseries{18.1812} & \bfseries{197.7314} \\
        \midrule
        \multicolumn{7}{l}{\emph{TE-388}} \\
        LATO.2~\cite{lato2}                   & 0.0991 & 0.2990 & 0.1509 & 0.2978 & 119.9100 & 931.6500 \\
        EdgeRunner~\cite{tang2024edgerunner} & 0.3990 & 0.7102 & 0.1261 & 0.1878 & 67.0844 & 651.5899 \\
        \shortname (Ours)                    & \bfseries{0.0415} & \bfseries{0.2075} & \bfseries{0.1829} & \bfseries{0.3106} & \bfseries{43.2544} & \bfseries{363.7326} \\
        \bottomrule
    \end{tabular}
\end{table}

\paragraph{Qualitative comparison.}
As illustrated in Figure.~\ref{fig:main_compare}, the two baselines exhibit characteristic failure modes.
EdgeRunner, an autoregressive generator, tends to output smoothed convex hulls with irregular triangulation and often drops thin or topologically isolated structures: limbs go missing, hair collapses into a single shell, and intricate facial features are washed out into approximate blobs.
LATO2, a diffusion generator over topology-preserving latents, recovers global shape well; in side-by-side comparison \shortname produces cleaner local triangulation and better preserves thin, branchy parts.
\shortname recovers these structures because (i) its cascaded occupancy vertex stage is sparse but not isotropic and naturally accommodates thin extrusions, and (ii) its closed-form connection head does not impose a manifold assumption, so two thin components that meet at a single vertex remain valid in the latent edge space. 
Similarly, as shown in Figure.~\ref{fig:second_compare}, we further compare \shortname against two leading proprietary commercial models. 
\shortname remains highly competitive with these commercial models in terms of overall geometry quality, fine-structure preservation, and topological fidelity.

\paragraph{Gallery for more results.}

\begin{figure}[!htbp]
    \centering
    \includegraphics[width=0.95\textwidth]{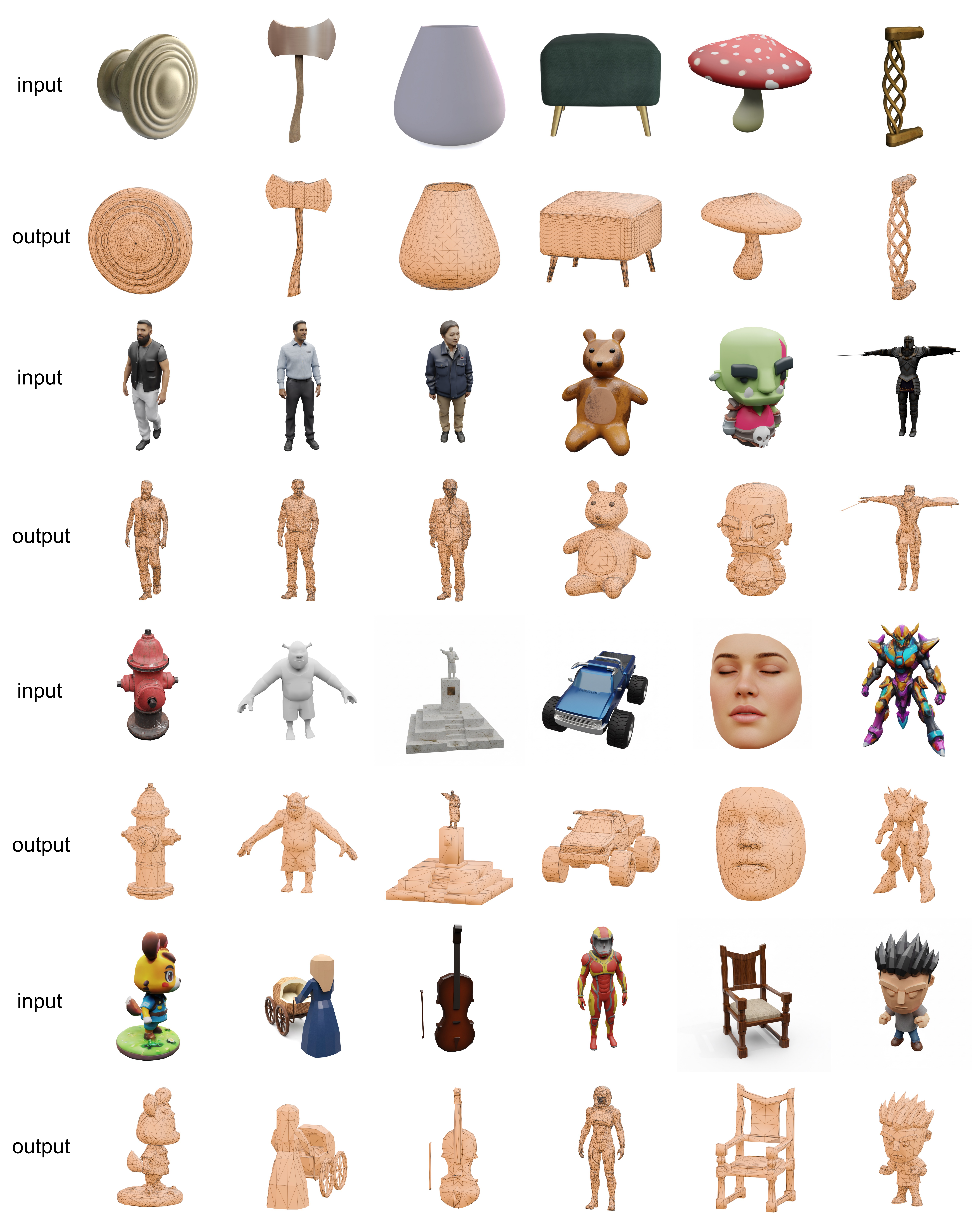}
    \caption{Additional image conditioned results. Our method generates meshes with both strong global structural stability and high-fidelity local details.\textit{Best viewed with zoom-in.}
    }
    \label{fig:gallery}
\end{figure}

Figure.~\ref{fig:gallery} presents additional image-conditioned generation results. It can be seen that our method yields meshes with strong global structural stability and fine geometric details, and generalizes effectively across a wide range of categories.

\begin{figure}[t]
    \centering
    \includegraphics[width=\textwidth]{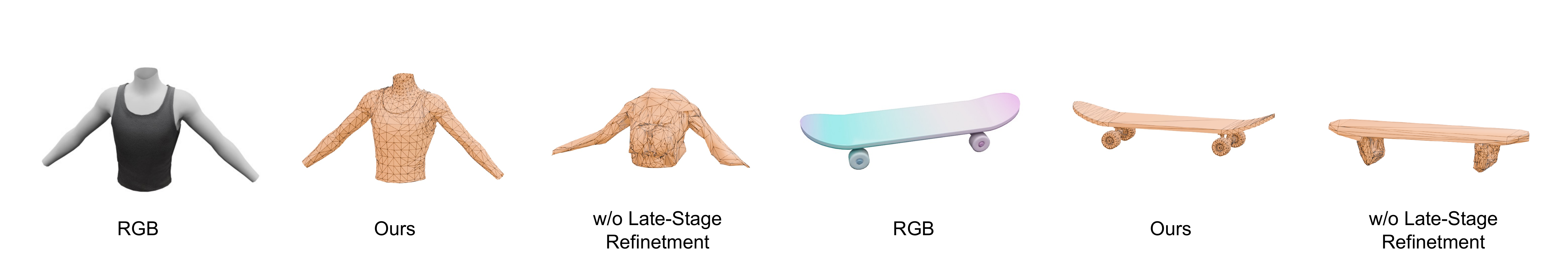}
    \caption{Qualitative ablation of late-stage progressive refinement, comparing continuous vertex generation with the late-stage refinement on structurally challenging examples.}
    \label{fig:ablation_compare}
\end{figure}

\subsection{Ablation Studies}
\label{subsec:exp_ablation}

We evaluate two design questions in our vertex generation pipeline.
The first study quantifies the effects of stage~2 level normalization and geometry-aware losses on TE-388.
The second study qualitatively examines whether combining stage~1 and stage~2 as late-stage progressive refinement improves structural stability.

\paragraph{Stage~2 normalisation and geometry-aware losses.}
We compare three matched stage~2 variants while keeping stage~1, stage~3, training data, and inference settings fixed.
The first uses level-wise Bernoulli normalization with only weighted BCE.
The second predicts unnormalized raw occupancy with geometry-aware losses added.
The full model combines normalization with all geometry-aware terms in Eq.~\eqref{eq:s2}.
As shown in Table.~\ref{tab:abl_s2}, model performance drops when the geometry-aware losses or the normalization strategy is removed.

\begin{table}[t]
    \centering
    \small
    \caption{Ablation of stage~2 normalisation and topology-aware losses.}
    \label{tab:abl_s2}
    \resizebox{\linewidth}{!}{%
    \begin{tabular}{lcccccc}
        \toprule
        Stage~2 variant & CD $\downarrow$ & HD $\downarrow$ & ULIP-I $\uparrow$ & Uni3D-I $\uparrow$ & FD-Incep. $\downarrow$ & FD-DINOv2 $\downarrow$ \\
        \midrule
        w/o. Geometry-aware Losses              & 0.0771 & 0.2734 & 0.1515 & 0.2501 & 52.6284 & 423.8286 \\
        w/o. Normalised Target       & 0.0576 & 0.2529 & 0.1647 & 0.2897 & 47.2389 & 390.2450 \\
        Ours & \bfseries{0.0415} & \bfseries{0.2075} & \bfseries{0.1829} & \bfseries{0.3106} & \bfseries{43.2544} & \bfseries{363.7326} \\
        \bottomrule
    \end{tabular}%
    }
\end{table}

\paragraph{Late-stage progressive refinement.}
We compare the full stage~1 to stage~2 cascade with a matched single-stage variant that attempts to generate the final $512^3$ vertex support directly.
As illustrated in Figure.~\ref{fig:ablation_compare}, replacing the late-stage refinement strategy with direct continuous upsampling degrades both the overall skateboard geometry and the fidelity of the wheel details.
The progressive model is expected to preserve the global layout established by stage~1 while stage~2 adds fine vertices within that support, yielding more stable structures than direct high-resolution generation.

%% file: sec/4_con.tex
\section{Conclusion}
We present a progressive native mesh generation framework developed as the foundation model of the Taobao~3D pipeline. In the future, \shortname can be applied to more general generation pipelines, such as serving as a component of large-scale multimodal models~\cite{openai2026gpt6}, generating 3D scenes based on scene graphs~\cite{SceneGraph}, or producing UV-space textures based on the unwrapping results \cite{SmartUV} of our methods.